\documentclass[10pt,twocolumn,letterpaper]{article}

\usepackage[pagenumbers]{wacv} %

\usepackage[table]{xcolor}
\usepackage{capt-of}   %
\definecolor{rankone}{RGB}{255, 210, 210}   %
\definecolor{ranktwo}{RGB}{255, 225, 190}   %
\definecolor{rankthree}{RGB}{255, 245, 200} %
\newcommand{\first}[1]{\cellcolor{rankone}\textbf{#1}}
\newcommand{\second}[1]{\cellcolor{ranktwo}#1}
\newcommand{\third}[1]{\cellcolor{rankthree}#1}

\usepackage{multirow}
\usepackage{makecell}
\usepackage{rotating}
\usepackage{pifont}
\newcommand{\cmark}{\textcolor{ForestGreen}{\ding{51}}}
\newcommand{\xmark}{\textcolor{BrickRed}{\ding{55}}}

\usepackage{pifont}
\usepackage[nointegrals]{wasysym} 
\newcommand{\parts}{\LEFTcircle}

\usepackage{graphicx} 
\usepackage{xcolor}

\usepackage{lipsum}
\usepackage{capt-of}

\definecolor{wacvblue}{rgb}{0.21,0.49,0.74}
\usepackage[pagebackref,breaklinks,colorlinks,allcolors=wacvblue]{hyperref}

\usepackage[normalem]{ulem}
\definecolor{rex_blue}{RGB}{0,0,255}

\definecolor{linli_green}{RGB}{0,155,0}

\def\ours{EVIS}

\title{EVIS: Real-Time Event Camera Simulation with Multimodal Supervision in NVIDIA Isaac Sim}

\author{Linli Shi
\quad Ruijun Zhang
\quad Ziyun Wang\thanks{Corresponding author.}\\
Johns Hopkins University\\
{\tt\small \{lshi42, rzhan158\}@jh.edu, \; claude.w@jhu.edu}
\vspace{-10mm}
}

\begin{document}
\twocolumn[{%
\renewcommand\twocolumn[1][]{#1}%
\maketitle
\begin{center}
  \centering
  \includegraphics[width=.95\textwidth]{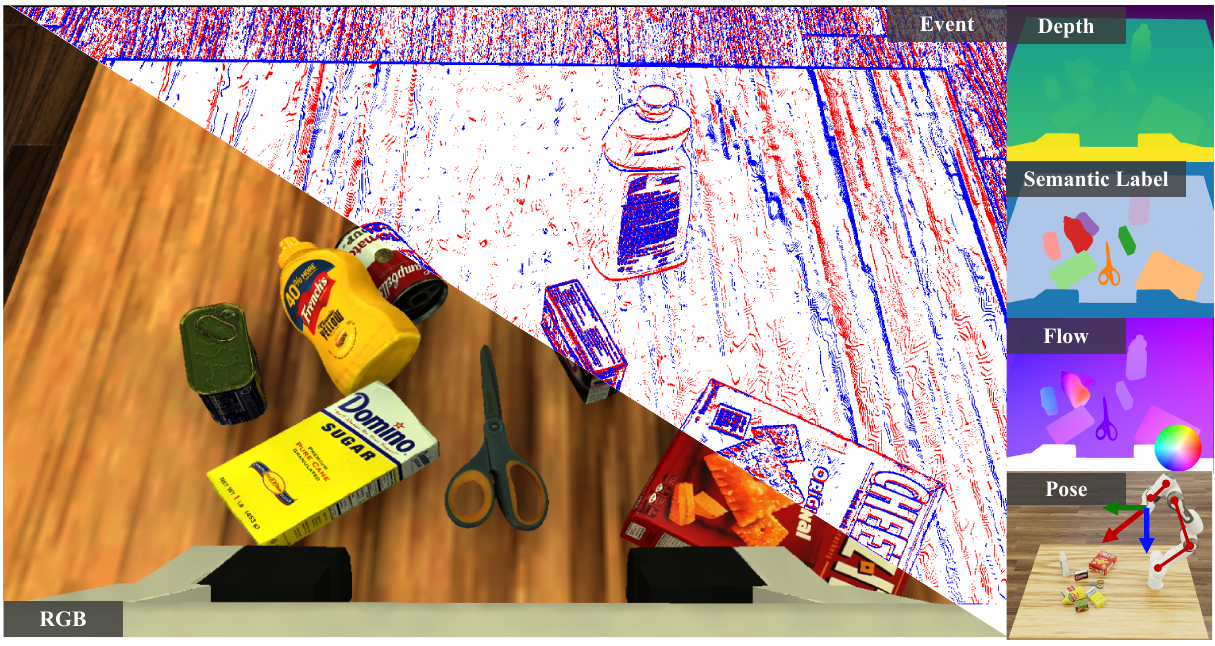}
  \vspace{-2mm}
  \captionof{figure}{EVIS is a physics-grounded event simulator capable of real-time event rendering for dynamic robot–scene interactions. By rendering directly from HDR log-intensity values in a physics engine, EVIS delivers high-fidelity event outputs paired with dense and continuous physical labeling.}
  \label{fig:teaser}
\end{center}%
}]
\begin{abstract}
Event cameras are increasingly adopted in embodied perception for their microsecond temporal resolution, high dynamic range, and resilience to motion blur. However, training event-based models for robotics requires large-scale, action-conditioned data with dense physical annotations that are difficult to collect in the real world. 

We introduce \textbf{EVIS}, an open-source physics-grounded event simulator integrated into NVIDIA Isaac Sim that generates events from linear-HDR radiance 
from a closed-loop robot training episode. Rather than relying on learning-based video interpolation or expensive dense rendering, EVIS exploits renderer-provided motion vectors and depth maps through bi-directional warping and depth-based splatting. This enables high throughput, real-time, and high-fidelity event generation. We evaluate EVIS across runtime efficiency, sim-to-real transfer, and zero-shot model compatibility. A rotation-speed estimator trained solely on EVIS events achieves 2.75\,rad/s MAE on real sensor data. Pretrained models for reconstruction, matching, and tracking perform competitively on EVIS events without any fine-tuning. EVIS sustains real-time generation across GPU-parallel environments on a single GPU. Code repository:
\url{https://github.com/spikelab-jhu/isaac-sim-event-camera-plugin}.
\end{abstract}

\section{Introduction}
\label{sec:intro}

Robots vision systems face significant challenges in conditions where conventional frame-based cameras struggle, such as rapid ego-motion, sudden illumination changes, and high-speed object motion. These conditions can produce motion blur and exposure artifacts that degrade standard image pipelines~\cite{gallego2020event}.
Event cameras offer a compelling alternative for these scenarios. With microsecond temporal resolution, asynchronous per-pixel brightness reporting, high dynamic range, and negligible motion blur, they preserve the visual information that matters most when traditional sensors fail.

Despite these advantages, real-world event data are scarce and difficult to annotate~\cite{perot2020learning}. Ground truth is rarely measured on the event stream itself: depth and pose typically come from companion sensors, and semantic labels from human annotators working on co-registered intensity frames. Both sources degrade precisely where event cameras are most useful. 
Under fast motion, LiDAR and RGB-D measurements suffer from motion distortion and limited temporal resolution, which require precise interpolation when they are used to label events. Intensity frames also suffer from motion blur when the event rate is high~\cite{zhang2014loam, pan2020high}.
A robot learner demands even more: \emph{action-conditioned} data with dense labels of physical and robot states, which allows it to reason about how observations would change under alternative actions rather than replaying a single recorded stream~\cite{levine2020offline}. Such annotation is infeasible to obtain in the real world, motivating the use of event simulators.

Existing event simulation approaches do not fully address the needs of event-based perception in robotics.
While video-based methods upsample low frame rate videos to synthesize photorealistic events~\cite{gehrig2020video, reda2022film}, 
the resulting events remain tied to prerecorded trajectories and therefore cannot simulate counterfactual outcomes under different robot actions or control policies.
Moreover, most source videos are recorded in low dynamic range, compressing the linear intensity changes that event cameras rely on. Their temporal resolution is also limited, and temporal super-resolution techniques often struggle with fast-moving objects~\cite{jiang2018super}. 
Finally, supervision is limited to annotations that already exist with the low frame rate recorded sequences.

Render-based methods~\cite{mueggler2017event} relax the fixed-recording constraints by allowing different trajectories under scripted scenes in the simulator. 
However, they still cannot provide interactive closed-loop feedback or dense physical annotations such as object poses, contact states, and velocities.
Physics-coupled simulators provide counterfactual rollouts and dense physical annotations. Yet, their renderers typically produce tone-mapped, low-dynamic-range imagery rather than the linear radiance, which creates a systematic photometric mismatch with the sensing model of event cameras, thereby introducing an irreducible sim-to-real gap.
To our knowledge, \textbf{no existing simulator jointly satisfies}: (a)~event generation on linear-HDR radiance, (b)~physically interactive environments with closed-loop action, (c)~synchronized access to dense geometric and physical supervision, and (d)~GPU-scale parallel rollouts that sustain real-time robot training.

We propose \textbf{EVIS}, a physics-grounded event data simulator enabling large-scale, real-time training with rich geometric and physical annotations. Built on NVIDIA Isaac Sim~\cite{mittal2023orbit}, EVIS leverages three key capabilities of modern physics-based simulators. First, the RTX ray-traced renderer produces scene-linear HDR imagery, allowing EVIS to evaluate its event model on physically faithful log-luminance rather than a tone-mapped surrogate. Second, Isaac Sim natively exposes per-pixel motion vectors and depth maps alongside each rendered frame at negligible additional cost. EVIS exploits these to synthesize intermediate frames via bidirectional motion-vector warping with depth-aware occlusion handling, achieving kHz-scale event generation without rendering every frame. Third, Isaac Sim steps multiple environments in parallel on a single GPU~\cite{makoviychuk2021isaac}. EVIS inherits this parallelism, sustaining real-time event generation across simultaneous robot training episodes while retaining synchronized access to depth, segmentation, poses, contact states, and other annotations.

Closing the sim-to-real gap is essential for large-scale event-based training~\cite{tan2025carladvs, stoffregen2020reducing}, yet event fidelity is difficult to measure directly since no ground-truth real event stream exists for a simulated scene. We therefore evaluate EVIS along three complementary axes.
For \textbf{sim-to-real transfer}, we train a rotation-speed estimator purely on EVIS events and test it on a real event camera, measuring whether synthetic training translates to real-sensor performance. For \textbf{zero-shot model compatibility}, we apply pretrained perception models, trained independently and never fine-tuned on EVIS data, directly to our simulated events across video reconstruction, wide-baseline feature matching, and point tracking, testing whether EVIS events are statistically close enough to real data for pretrained models to perform competitively. For \textbf{runtime efficiency}, we benchmark generation throughput under varying warp factors and GPU-parallel configurations. Experiments demonstrate that EVIS generates realistic events that effectively support both training and zero-shot inference while sustaining real-time throughput on a single GPU. Our contributions can be summarized as follows.
\begin{itemize}
    \item We introduce \ours{}, an open-source {physics-grounded} event data simulator that enables large-scale
    generation of action-conditioned event data alongside rich geometry and physics annotations. 
    \item We propose a bidirectional warping method that utilizes motion vectors and depth maps in the rendering pipeline to improve render efficiency, enabling high-temporal-resolution, real-time event data simulation.

    \item We comprehensively benchmark \ours{} in terms of runtime efficiency, sim-to-real fidelity, zero-shot downstream performance including \emph{video reconstruction, point tracking,  and feature matching}. Experiments show that \ours{} generate realistic events effectively support large-scale training.
    
\end{itemize}

\section{Related Work}
\subsection{Physics-Based Simulation for Embodied Vision}
\label{sec:physicssim}
Vision for robotics relies on physics-based simulation for safe, reproducible interaction data. Such simulators provide a closed perception-action loop and synchronized ground truth from multiple sensors.
General-purpose engines such as Gazebo~\cite{koenig2004design} and MuJoCo~\cite{todorov2012mujoco} excel at contact-rich manipulation and locomotion, while CARLA~\cite{dosovitskiy2017carla} and AirSim~\cite{shah2017airsim} layer photorealistic sensing atop Unreal Engine. However, these renderers usually produce tone-mapped, low-dynamic-range images. Even MuJoCo, whose physics scales to thousands of environments via MJX, is limited to LDR rendering. 
These simulators typically produce tone-mapped images for visual rendering, which may compress the scene's intensity gradients and widen the sim-to-real gap for event cameras that operate on linear irradiance. NVIDIA Isaac Sim~\cite{mittal2023orbit} is a widely adopted robotics simulation platform that pairs GPU-accelerated PhysX~5 with an RTX ray-traced renderer capable of producing scene-linear HDR imagery. It steps thousands of environments in parallel on a single GPU~\cite{makoviychuk2021isaac} and natively exposes per-pixel motion vectors alongside synchronized ground truth. EVIS builds on Isaac Sim to bring these capabilities to event camera simulation.

\begin{table*}[t]
  \centering
  \setlength{\tabcolsep}{9pt}
  \renewcommand{\arraystretch}{1}
  \caption{Comparison of event-camera simulators.
  \cmark~yes \quad \parts~partial \quad \xmark~no \quad --~not applicable.
  \emph{Action}: a changed input drives a different physical evolution;
  \emph{Event-in-loop}: online closed-loop rollout;
  \emph{Linear intensity}: events on linear radiance, not tone-mapped LDR;
  \emph{Temporal res.}: sub-frame event timing is recovered;
  \emph{Multi-env Parallelism}: multiple environments run in parallel on GPU;
  \emph{Annotation}: synchronized geometry and physical ground truth.}
  \begin{tabular}{@{}c l c c c c c c@{}}
  \toprule
  & Method & Action & \makecell{Event-\\in-loop} & \makecell{Linear\\intensity}
    & \makecell{High Temporal\\ Resolution} & \makecell{Multi-env\\Parallelism} & \makecell{Physics\\ Annotation} \\
  \midrule
  \multirow{3}{*}{\rotatebox[origin=c]{90}{\shortstack{Frame-\\Based}}}
    & v2e~\cite{hu2021v2e}                    & -- & -- & \xmark & \cmark & -- & -- \\
    & V2CE~\cite{zhang2024v2ce}                  & -- & -- & \xmark & \cmark & -- & -- \\
    & DVS-Voltmeter~\cite{lin2022dvs} & -- & -- & \xmark & \cmark & -- & -- \\
  \midrule
  \multirow{3}{*}{\rotatebox[origin=c]{90}{\shortstack{Render-\\based}}}
    & ESIM~\cite{rebecq2018esim}          & \parts & -- & \parts & \cmark & \parts & \parts \\
    & Mueggler~\etal~\cite{mueggler2017event} & \parts & -- & \cmark & \parts & \parts & \parts \\
    & EvDNeRF~\cite{bhattacharya2024evdnerf}    & \parts & -- & \parts & \parts & \parts & \parts \\
  \midrule
  \multirow{4}{*}{\rotatebox[origin=c]{90}{\shortstack{Physics-\\based}}}
    & Gazebo~\cite{kaiser2016towards} & \cmark & \cmark & \xmark & \parts & \parts & \cmark \\
    & AirSim~\cite{shah2017airsim}    & \cmark & \cmark & \xmark & \parts & \parts & \cmark \\
    & CARLA~\cite{dosovitskiy2017carla}      & \cmark & \cmark & \xmark & \parts & \parts & \cmark \\
    & \textbf{EVIS (ours)}    & \cmark & \cmark & \cmark & \cmark & \cmark & \cmark \\
  \bottomrule
  \end{tabular}
  
  \label{tab:sim_comparison}
\end{table*}

\subsection{Event Camera Simulation}
\label{sec:eventsim}
Event simulators replace expensive, hard-to-label real recordings with large-scale synthetic data carrying synchronized ground truth~\cite{gallego2020event}. Existing methods can be categorized into three families, as summarized in Table~\ref{tab:sim_comparison}. 

\noindent\textbf{Frame-based.} Early work converts existing video into events by frame differencing~\cite{katz2012live, bi2017pix2nvs}, while Vid2E upsamples with Super-SloMo before applying the ESIM model to reach DVS rates~\cite{gehrig2020video, jiang2018super, rebecq2018esim}. A parallel line focuses on sensor non-idealities~\cite{jiang2024adv2e}: v2e introduces pixel noise~\cite{hu2021v2e}, DVS-Voltmeter employs stochastic per-pixel dynamics~\cite{lin2022dvs}, and others contribute calibrated~\cite{10.3389/fnins.2021.702765} or differentiable~\cite{greene2025pytorch} forward models. Learning-based methods bypass explicit modeling and map video directly to event voxels~\cite{zhang2024v2ce, lou2025v2v, zhu2021eventgan}. In most cases, annotations are inherited from the source video.

\noindent\textbf{Rendering-based.} Rendering-based simulators generate events from a 3D scene. Unlike video-based methods, they can independently control scene geometry, lighting, and camera trajectory. Mueggler~\etal render scripted Blender trajectories~\cite{mueggler2017event}, and ESIM couples an OpenGL/Unreal backend with motion-adaptive sampling~\cite{rebecq2018esim}. Recent work advances photometric realism through physically based~\cite{han2024physical, manabe2024monte} or path-traced~\cite{li2026eventtracer} rendering, while neural approaches sample events from reconstructed radiance fields~\cite{bhattacharya2024evdnerf}. These methods offer full control over scene, lighting, and trajectory, yet without a physics engine they can neither close a perception-action loop nor provide annotations beyond depth and flow.

\noindent\textbf{Physics simulation-based.} The third family generates events inside a physics simulator, placing them in a closed perception-action loop alongside other sensors. The Gazebo DVS plugin differences grayscale frames~\cite{kaiser2016towards}, while AirSim and CARLA apply ESIM atop Unreal Engine~\cite{shah2017airsim, dosovitskiy2017carla}. However, these platforms
typically produce tone-mapped LDR frame, which may compromise event fidelity for sensors that operate on linear irradiance. EVIS instead operates on Isaac Sim's linear-HDR buffer across GPU-parallel environments and recovers sub-frame timing through motion-vector warping, producing high-temporal-resolution events in real time.

\subsection{Sim-to-Real Transfer for Event Vision}
\label{sec:sim2realtrans}

The sim-to-real gap is a well-known challenge across robotics and computer vision~\cite{zhao2020sim} especially event-based vision. Prior work has measured this gap across tasks including optical flow~\cite{luo2023mdrflow}, depth estimation~\cite{bartolomei2025depthanyevent}, hand pose estimation~\cite{rudnev2021eventhands}, and object detection~\cite{tan2025carladvs}. The gap is not negligible: a detector trained on a crude simulator's events reaches only approximately 7 percent of the performance achieved by one trained on faithfully simulated events~\cite{zhu2021eventgan}, and aligning simulator statistics to the target sensor improves reconstruction by 20--40 percent~\cite{stoffregen2020reducing}. A common strategy to mitigate this gap is simulation-to-real fine-tuning; for instance, a recognizer improves from 75 to 81 percent accuracy with domain randomization alone, and further approaches real-data-trained performance after light fine-tuning~\cite{gehrig2020video}, implicitly confirming that the gap exists but can be reduced with sufficient simulator fidelity. Inspired by prior work, we evaluate pretrained model performance and cross-validate on synthetic and real data to quantify the sim-to-real gap, establishing direct comparisons with prior event simulators.

\section{Method}
\label{sec:method}

\subsection{Problem Formulation}
\label{sec:problem}

An event camera independently monitors the log-intensity at each pixel and asynchronously emits an event $(x, y, t, p)$, encoding pixel location, timestamp, and polarity, whenever the brightness change exceeds a contrast threshold. Given world state $\mathcal{S}_t$ and robot action $\mathbf{a}_t$, the physics engine evolves the scene as
\begin{equation}
    \mathcal{S}_{t+\Delta t} = \mathcal{P}(\mathcal{S}_t, \mathbf{a}_t),
\end{equation}
where $\mathcal{P}$ is the physical transition function and $\mathbf{a}_t$ displaces both the robot and environment. Rendering the state can be considered as sampling the linear-HDR color at pixel $\mathbf{u}$, \begin{equation} I(\mathbf{u},t)=\mathcal{I}(\mathcal{S}_t,\mathbf{u}), \end{equation}
From action conditioned world state, EVIS generates a synchronized event stream $\mathcal{E}_{[t,t+\Delta t]}=\{(x_i, y_i, t_i, p_i)\}_i$ and task supervision $\mathcal{Y}_{t+\Delta t}$, which can include depth, optical flow, pose, segmentation, contact states. 

\subsection{HDR Event Formation Model}
\label{sec:eventmodel}
We synthesize events from the linear radiance produced by the RTX renderer, prior to tone mapping or gamma correction. At pixel $\mathbf{u}=(x,y)$, we first convert the rendered color to luminance using Rec.~709 weights:
\begin{equation}
I = 0.2126\,R + 0.7152\,G + 0.0722\,B.
\end{equation}
This yields a log-intensity $L(\mathbf{u},t)=\log\!\big(I(\mathbf{u},t)+\epsilon\big)$. Replacing $I$ with individual R, G, B channels and applying the event model to each independently produces color events. Each pixel maintains a reference level $L_{\mathrm{ref}}(\mathbf{u})$ and fires whenever $L$ departs from $L_{\mathrm{ref}}$ by the contrast threshold $C$. Over a discrete rendering step, the departure $\Delta L = L - L_{\mathrm{ref}}$ may cross multiple thresholds, so the pixel emits
\begin{equation}
n(\mathbf{u}) = \big\lfloor |\Delta L| / C \big\rfloor
\quad\text{events of polarity}\quad
p = \operatorname{sgn}(\Delta L),
\end{equation}
and its reference is updated as $L_{\mathrm{ref}} \leftarrow L_{\mathrm{ref}} + p\,n\,C$, retaining the sub-threshold residual for the next step. Operating on linear-HDR radiance ensures that 
the accumulator recovers the correct event count even under the large per-step contrast that fast motion induces.

\begin{figure*}[t]
\centering
\includegraphics[width=\linewidth]{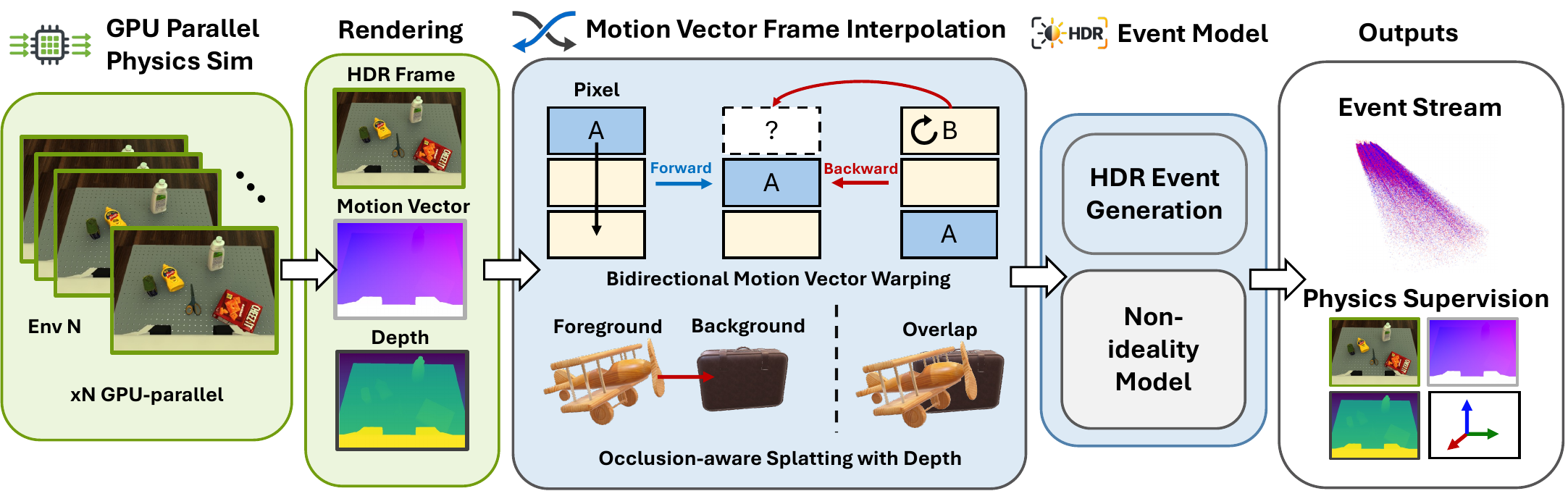}
\caption{\textbf{Overview of the EVIS pipeline.} A physics engine advances the world state under robot actions, and the renderer produces linear-HDR keyframes along with per-pixel motion-vector and depth maps. HDR frames are then interpolated to a kHz-scale stream via bidirectional motion-vector warping with occlusion-aware depth splatting. The resulting HDR frames are fed into an event generation model that applies configurable sensor non-idealities to produce realistic events. Physics supervision is extracted from the same simulation process, guaranteeing events and labels synchronization by construction.}
\label{fig:pipeline}
\end{figure*}

\subsection{Bidirectional Motion Vector Warping}
\label{sec:warp}
Rendering and physics simulation dominate the pipeline's cost, yet a faithful event stream requires a kHz-scale frame rate, and closed-loop robot training requires it in real time. We therefore render keyframes at a low rate $f_k$ and synthesize the $K{-}1$ intermediate frames from renderer-provided motion vectors and depth maps, giving the event model an effective rate of $f_k K$ while keeping rendering real-time.

At each keyframe, the RTX renderer produces a color frame alongside a motion-vector map and a depth map at negligible additional cost (\cref{tab:benchmark}, row $1000{\times}1$ vs.\ $1000{\times}1$+). The renderer provides backward-pointing motion vectors: each pixel in keyframe $B$ stores a displacement to its corresponding position in the previous keyframe $A$. The field $\mathbf{m}_B$ thus encodes the exact $A{\to}B$ correspondence. For keyframe $A$, we approximately reuse its own backward motion $\mathbf{m}_A$ as a forward velocity estimate. Assuming linear motion across the keyframe gap, we synthesize the intermediate frame at fraction $f = i/K$, $i \in \{1,\dots,K{-}1\}$, by splatting $A$ forward and $B$ backward to the target positions
\begin{equation}
\mathbf{q}_A(\mathbf{p}) = \mathbf{p} - f\,\mathbf{m}_A(\mathbf{p}),
\qquad
\mathbf{q}_B(\mathbf{p}) = \mathbf{p} + (1{-}f)\,\mathbf{m}_B(\mathbf{p}).
\label{eq:warp-targets}
\end{equation}
These target values are then bilinearly interpolated to integer pixels.
When multiple sources land on the same target, we resolve the occlusion with a per-pixel $z$-buffer that retains only the nearest surface:
\begin{equation}
\widehat{I}(\mathbf{q}) = I_{X^\star}(\mathbf{p}^\star), \qquad
(X^\star, \mathbf{p}^\star) = \operatorname*{arg\,min}_{(X,\mathbf{p}) \in \mathcal{S}(\mathbf{q})} \widetilde{Z}(\mathbf{p}),
\label{eq:zbuffer-splat}
\end{equation}
where $\mathcal{S}(\mathbf{q})$ collects all sources whose bilinear footprint covers $\mathbf{q}$. The intermediate depth $\widetilde{Z}$ is linearly interpolated from the two rendered keyframe depths along the correspondence:
\begin{equation}
\widetilde{Z}(\mathbf{p}) = (1{-}f)\,Z_A(\mathbf{p}_A) + f\,Z_B(\mathbf{p}_B),
\label{eq:depth-interp}
\end{equation}
with $\mathbf{p}_A, \mathbf{p}_B$ the endpoints in $A$ and $B$.

The backward-warped $B$ serves as the primary result, since $\mathbf{m}_B$ provides the exact $A{\to}B$ correspondence. The forward-warped $A$ fills only $B$'s disocclusion holes.
All intermediates across environments and cameras are batched into a single scatter call per warp direction, keeping the interpolation stage GPU-efficient.

Since anti-aliasing assigns silhouette-fringe pixels the background's near-zero motion, leaving a ghost outline under the warp, we disable it during interpolation. 
We provide an optional over-rendering margin: rendering extra pixels on each side and cropping them off to solve the artifacts near the image boundary.

\definecolor{oursrow}{gray}{0.93}
\begin{table*}[t]
\centering
\renewcommand{\arraystretch}{1.15}
\setlength{\tabcolsep}{5pt}
\caption{\textbf{Simulation method vs.\ downstream task quality.} We test the events simulated by different methods on the reconstruction task and the feature matching task. Configs: \emph{keyframe rate}${\times}$\emph{upsampling factor}. \emph{Left:} E2VID\cite{rebecq2019high} reconstruction evaluated against ground-truth
luminance (CLAHE-normalized). \emph{Right:} event-to-event feature matching using Match-Any-Events\cite{zhang2026match}. \emph{Time:} The actual time it takes to generate a 30s event sequence (excluding I/O) for $30\times8$ config. RTF (real-time factor) is the ratio of event-time to wall-clock time; RTF$\,{\geq}\,1$ means the simulator produces events at least as fast as they would occur physically. } 
\footnotesize
\begin{tabular}{l l @{\hspace{12pt}} ccc @{\hspace{12pt}} cccc @{\hspace{12pt}} cc}
\toprule
\multirow{2}{*}{\textbf{Method}} & \multirow{2}{*}{\textbf{Config}}
& \multicolumn{3}{c}{\textbf{Reconstruction}} & \multicolumn{4}{c}{\textbf{Feature matching}}
& \multicolumn{2}{c}{\textbf{Efficiency ($30\times8$)}}\\
\cmidrule(lr){3-5}\cmidrule(lr){6-9}\cmidrule(lr){10-11}
& & SSIM~$\uparrow$ & MSE~$\downarrow$ & LPIPS~$\downarrow$
& Prec~$\uparrow$ & AUC@$5^\circ$~$\uparrow$ & AUC@$10^\circ$~$\uparrow$ & AUC@$20^\circ$~$\uparrow$
& Time~$\downarrow$ & RTF~$\uparrow$\\
\midrule
\rowcolor{oursrow}
& $1000{\times}1$ & {\textbf{0.579}} & 0.079 & 0.279 & 73.00 & 0.437 & 0.573 & 0.686 & & \\
\rowcolor{oursrow}
& $250{\times}4$ & 0.575 & 0.074 & {\textbf{0.275}} & \textbf{74.87} & \textbf{0.455} & \textbf{0.605} & \textbf{0.737} & & \\
\rowcolor{oursrow}
& $125{\times}8$
& {0.533} & {0.073} & {0.279} & {72.08} & {0.421} & {0.566} & {0.697} & & \\
\rowcolor{oursrow}
\multirow{-3}{*}{\textbf{EVIS}} & $60{\times}4$ & 0.496 & \textbf{0.055} & 0.278 &72.86 &0.412 &0.586 &0.725 & & \\
\rowcolor{oursrow}
& $30{\times}8$ & {0.431} & 0.058 & {0.305} &70.39 &0.411 &0.563 &0.690
& \cellcolor{oursrow}\multirow{-5}{*}{\textbf{25}} & \cellcolor{oursrow}\multirow{-5}{*}{\textbf{1.17$\times$}} \\
\midrule
\multirow{2}{*}{Vid2E (SSM)~\cite{gehrig2020video}}
& $125{\times}8$ & 0.531 & 0.064 & 0.284 & 69.41 & 0.322 & 0.469 & 0.585 & & \\
& $30{\times}8$ & 0.428 &0.077 &0.355 &71.41 &0.402 &0.540 &0.661
& \multirow{-2}{*}{42} & \multirow{-2}{*}{0.71$\times$} \\
\midrule
\multirow{2}{*}{FILM~\cite{reda2022film}}
& $125{\times}8$ &0.518 &0.087 &0.314 &70.20 &0.332 &0.465 &0.578 & & \\
& $30{\times}8$ &0.394 &0.093 &0.350 &73.31 &0.437 &0.580 &0.690
& \multirow{-2}{*}{252} & \multirow{-2}{*}{0.12$\times$} \\
\midrule
\multirow{2}{*}{RIFE~\cite{huang2022real}}
& $125{\times}8$ &0.530 &0.072 &0.289 &69.42 &0.310 &0.433 &0.538 & & \\
& $30{\times}8$ &0.426 &0.077 &0.342 &71.26 &0.415 &0.547 &0.658
& \multirow{-2}{*}{42} & \multirow{-2}{*}{0.71$\times$} \\
\midrule
\multirow{2}{*}{GIMM-VFI-R~\cite{guo2024generalizable}}
& $125{\times}8$ &0.521 &0.093 &0.315 &69.32 &0.338 &0.450 &0.547 & & \\
& $30{\times}8$ &0.401 &0.087 &0.341 &72.32 &0.436 &0.573 &0.686
& \multirow{-2}{*}{902} & \multirow{-2}{*}{0.06$\times$} \\
\bottomrule
\end{tabular}

\label{tab:maintable}
\end{table*}

\begin{figure*}[t]
\centering
\includegraphics[width=\linewidth]{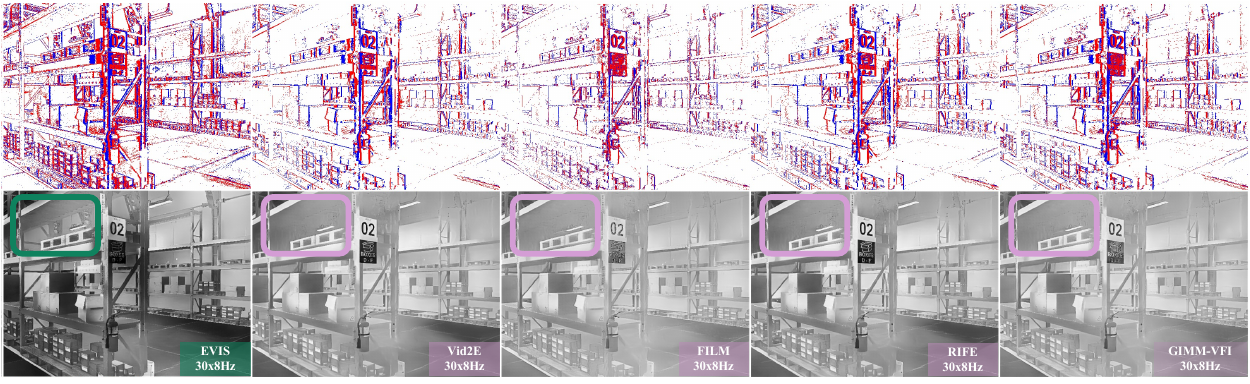}
\caption{\textbf{Video reconstruction with pretrained E2VID~\cite{rebecq2019high}.} Top: $10$\,ms event accumulations (red: ON, blue: OFF) from each configuration. Bottom: corresponding E2VID reconstruction results. }
\label{fig:downstream}
\end{figure*}

\subsection{Sensor Calibration and Non-idealities}
\label{sec:noise}
Inspired by prior work on event camera noise model
~\cite{hu2021v2e, rebecq2018esim, gallego2020event}, EVIS provides six optional, per-camera configurable non-idealities to further close the sim-to-real gap:
\begin{itemize}
  \item \textbf{Threshold mismatch.} The scalar threshold $C$ is replaced by per-pixel ON/OFF thresholds
  \begin{equation}
  \theta^{\text{ON}},\; \theta^{\text{OFF}} \sim
  \max\!\big(\mathcal{N}(C,\, \sigma_\theta^2),\, \theta_{\min}\big),
  \end{equation}
  sampled once per pixel to model the fixed-pattern mismatch of a real sensor array.

  \item \textbf{Refractory period.} After firing at $t_{\text{last}}$, a pixel suppresses subsequent events until $t - t_{\text{last}} \ge \tau_r$.

  \item \textbf{Leak events.} Junction leakage produces spurious ON events. In each synthesis step of duration $\Delta t$, a pixel leaks with probability
  \begin{equation}
  P_{\text{leak}} = r_{\ell}\,\Delta t.
  \end{equation}

  \item \textbf{Shot noise.} Random-polarity noise events whose rate increases with darkness:
  \begin{equation}
  P_{\text{shot}} = r_{s}\,\Delta t\,\big(1 - \bar{I}\big),
  \end{equation}
  where $\bar{I} \in [0,1]$ is the normalized intensity and polarity is drawn uniformly.

  \item \textbf{Hot pixels.} A fixed random subset $\mathcal{H}$ of pixels fires at a persistently high rate: $P_{\text{hot}} = r_{h}\,\Delta t$ for $\mathbf{p} \in \mathcal{H}$.

    \item \textbf{Defocus.} Prior to the event generation, the linear intensity is convolved with a separable Gaussian kernel.

\end{itemize}

\label{sec:isaacsim}

\section{Experiments}
\label{sec:exp}

\subsection{Zero-Shot Model Generalization}
\label{sec:downstream}
Here we probe whether perception models, trained on real datasets, or on other simulation data, generalize on \ours{}. We evaluate three diverse tasks: video reconstruction, wide-baseline feature matching, and point tracking, each exercising a different aspect of event fidelity from photometric consistency to geometric accuracy.
\paragraph{Video Reconstruction}
We reconstruct an intensity video from the events with the E2VID~\cite{rebecq2019high}, windowed at the ground-truth timestamps ($50$\,Hz), and score each frame against the ground-truth luminance with three standard metrics: the mean squared error (MSE),
structural similarity~\cite{wang2004ssim}, 
and LPIPS~\cite{zhang2018unreasonable}.
Since events encode only intensity \emph{changes}, the absolute scale and offset of the reconstruction are unknown; following the evaluation protocol of E2VID~\cite{rebecq2019high}, we histogram-equalize both frames before scoring (CLAHE) so the metrics measure structure rather than global tone. 

 We compare against Vid2E~\cite{gehrig2020video} and three representative frame-interpolation methods: FILM\cite{reda2022film}, RIFE\cite{huang2022real}, and GIMM-VFI\cite{guo2024generalizable}. Each followed by the same ESIM backend. Notably, these interpolation networks are mostly trained on LDR frames and cannot directly transfer to HDR inputs; we therefore tone-map
to LDR before interpolation.
For SSIM, EVIS at the $1000\times1$ configuration achieves the highest score of $0.579$. For LPIPS, EVIS at $250\times4$ attains the best value of $0.275$, as shown in~\cref{tab:maintable}. 
When upsampled to 1000Hz, \ours{} produces similar results and outperforms prior work in reconstruction fidelity. 
Even under identical settings such as $125\times8$ or $30\times8$, EVIS surpasses all baselines, underscoring the high fidelity of our event simulation. We visualize the reconstructed frames alongside the corresponding event representations. EVIS exhibits sharper structural detail and fewer artifacts (highlighted in~\cref{fig:downstream}), indicating that our simulator produces events whose statistics approximate the real-data distribution.

\paragraph{Wide Baseline Matching}
We construct an indoor scene and sampled viewpoints on trajectories using identical method in Match-Any-Events~\cite{zhang2026match} and match events across the two arbitrary views. We evaluate following the same protocol and report matching precision, defined as the ratio of correct matches (epipolar distance below $1\times10^{-4}$) to total matches, and the AUC of the pose error at $5/10/20^\circ$. 

\begin{figure}[t]
\centering
\includegraphics[width=\linewidth]{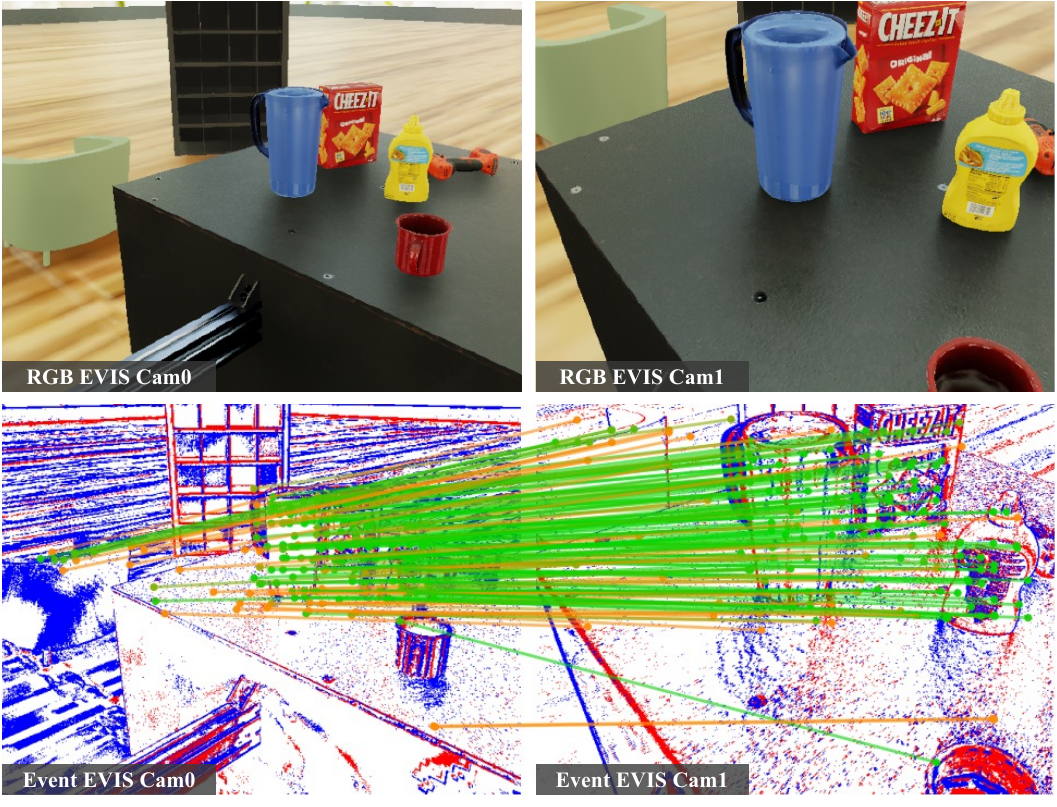}
\caption{\textbf{Event-to-event feature matching.}
Top: the RGB view from each of the two cameras.
Bottom: the event streams of the two views(ON red, OFF blue) with the
correspondences found by the  matcher drawn as colored lines. Green: low epipolar error. Orange: high epipolar error.}
\label{fig:match}
\end{figure}

Across all configurations, EVIS yields accuracy consistent with the original result reported in the paper~\cite{zhang2026match}, with the $250\times4$ setting achieving the best matching precision. ~\cref{fig:match} visualizes a matching result at $1000$\,Hz. EVIS achieves performance comparable to the baselines on the Match-Any-Events model, though it is marginally surpassed (by roughly 2\% in precision) under the $30\times8$ setting. We attribute this to the fact that Match-Any-Events is trained on events generated by ESIM from LDR frames~\cite{zhang2026match}. Moreover, at a $30$\,Hz rendering rate the linear motion assumption underlying our motion-vector warping is more frequently violated, which favours baseline methods that rely on learned frame interpolation to produce denser sub-frames. EVIS remains highly competitive in terms of efficiency: it is the only method capable of generating events in real time under the $30\times8$ configuration.

\begin{figure}[t]
\centering
\includegraphics[width=\linewidth]{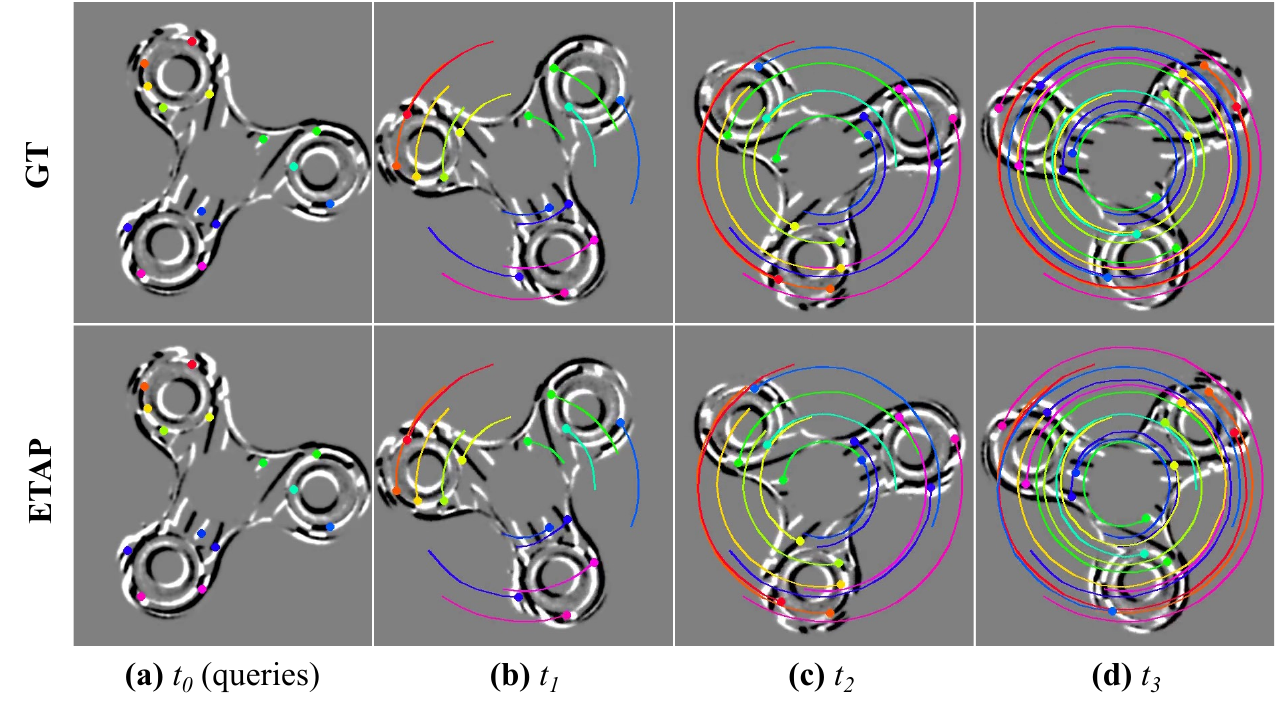}
\caption{\textbf{Point tracking with  ETAP~\cite{hamann2025etap} on EVIS-simulated events.} A fidget spinner rotates at high speed. Query points are initialized at $t_0$\,(a) and tracked through subsequent frames (b--d). Top: ground-truth trajectories. Bottom: ETAP predictions. 
}
\label{fig:etap}
\end{figure}

\paragraph{Point Tracking}
We evaluate event-based point tracking on a fidget spinner at high speed following~\cite{hamann2025etap}.
We sample query points on the first frame and track them through EVIS's simulated 1000Hz events with  ETAP~\cite{hamann2025etap}.
Colored curves show the ETAP point tracking trajectories under 300Hz. 
As shown in \cref{fig:etap}, ETAP successfully recovers the circular motion pattern, with predicted trajectories closely matching the ground truth throughout the sequence, demonstrating the data coherency.

\vspace{-2mm}

\begin{figure}[htbp]
\centering
\includegraphics[width=\linewidth]{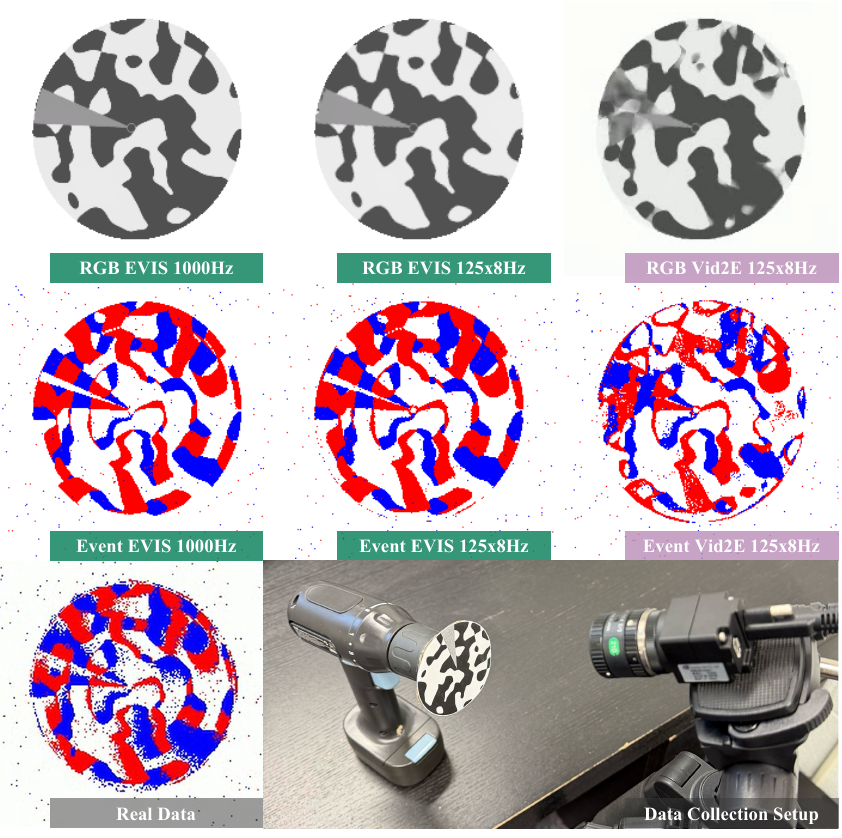}
\caption{\textbf{Spinning textured disk at 50\,rad/s.} Top row: interpolated RGB frames. Middle row: 5\,ms event windows. Bottom left: real events from an EvC3A camera. Bottom right: data collection setup. 
Super-SloMo used by Vid2E~\cite{gehrig2020video} produces blurry and misaligned texture under the high speed rotation.}
\label{fig:sim2real}
\end{figure}

\begin{table}[h]
\centering
\renewcommand{\arraystretch}{1.2}
\setlength{\tabcolsep}{8pt}
\caption{\textbf{Sim-to-real rotational speed estimation.} Each model is trained solely on EVIS or Vid2E synthetic events and tested on real clips. Gen.\ time is the compute-only wall time without I/O.
The top, second, and third performance are highlighted in \colorbox{rankone}{red}, \colorbox{ranktwo}{orange}, and \colorbox{rankthree}{yellow}, respectively.}
\footnotesize
\begin{tabular}{l l ccc}
\toprule
\textbf{method} & \textbf{config}
& \textbf{MAE}~(rad/s)~$\downarrow$ & \textbf{Var}~$\downarrow$ & \textbf{Gen.\ time}~$\downarrow$\\
\midrule
\multirow{3}{*}{EVIS}
      & $1000{\times}1$ & \first{2.75}  & \second{7.79}  & {$5{:}32$}\\
      & $250{\times}4$  & \second{5.75} & \first{7.53}   & \second{2{:}35}\\
      & $125{\times}8$  & \third{7.10}  & \third{14.35}  & \first{1{:}50}\\
\cmidrule(lr){1-5}
Vid2E & $125{\times}8$  & $8.25$        & $57.19$        & \third{$4{:}26$}\\
\bottomrule
\end{tabular}
\end{table}

\subsection{Sim-to-Real Rotational Speed Estimation}
\label{sec:sim2real}
The direct test of simulator fidelity is whether a model trained entirely on synthetic events transfers to a real sensor without fine-tuning. Inspired by event-based spin estimation deployed in competitive robotic table tennis~\cite{durr2026outplaying}, we design a sim-to-real benchmark for rotational speed estimation of a fast-spinning textured disk. 

\textbf{Data generation} To generate training data, we randomize rotation speed ($[15, 70]$\,rad/s), disk tilt ($[25, 45]^\circ$), disk center (within one quarter of the frame), camera standoff ($[0.8, 1.1]$\,m), and per-clip defocus ($0.1$--$0.5$\,px Gaussian blur). We produce 1500 clips per texture (7500 total), each 0.5\,s at $640{\times}480$, in three variants: no warping, $4{\times}$, and $8{\times}$ warp interpolation. \textbf{Vid2E baseline.} We upsample the same $125$\,Hz rendered keyframes $8{\times}$ with Super-SloMo and generate events at the same contrast threshold. \textbf{Speed Estimator.} A per-polarity, 128-bin event voxel (256 channels) is fed to a ResNet-18~\cite{he2016deep} with its first convolution widened to 256 input channels; a single linear head regresses rotation speed under MSE loss. \textbf{Real data.} We capture the real disk with a CenturyArks SilkyEvCam (EvC3A). We exploit the rotational periodicity of the event stream: we bin events into $1$\,ms frames and
annotate the ground-truth rotation speed by finding the dominant FFT peak. 
\textbf{Evaluation.} We report mean average error of rotational speed in rad/s and variance.

Across the evaluated settings, EVIS configurations exhibit lower MAE and lower variance compared to Vid2E. The full-rate EVIS 1000×1 achieves the lowest overall MAE (2.75 rad/s). Notably, the EVIS 125×8 configuration achieves lower error than Vid2E 125×8 while running 2.4× faster. Regarding output consistency, Vid2E exhibits a variance of 57.19, approximately 4× that of the highest EVIS configuration. We provide qualitative context for this discrepancy in~\cref{fig:sim2real}. At 50 rad/s, the disk rotates $\sim23^\circ$ 
between 125 Hz keyframes. Vid2E's underlying Super-SloMo framework struggles to compute reliable optical flow on repetitive textures, yielding frame artifacts that propagate into the rendered event stream. Visual comparison with real event captures from the CenturyArks sensor (bottom left) shows closer alignment with EVIS outputs than Vid2E.

\subsection{Scalability and Throughput}
\label{sec:benchmark}
 We measure throughput by the real-time factor (RTF), defined as the ratio of generated event-time to wall-clock time. We report compute-only generation time (physics, rendering, interpolation, and event model), excluding one-time Isaac Sim startup and disk I/O, as shown in~\cref{tab:benchmark}. All measurements use a single RTX 5090 on the warehouse scene, a medium-complexity environment.

At a $1000$\,Hz effective rate, EVIS $125{\times}8$ generates $10$\,s of event data in $25.0$\,s, $2.5{\times}$ faster than Vid2E $125{\times}8$, because motion-vector warping reuses renderer byproducts while Vid2E runs interpolation on every keyframe gap. 
Comparing EVIS $1000{\times}1$ with $1000{\times}1$+ (which additionally renders motion vectors and depth maps), the overhead is modest (RTF $0.12$ vs.\ $0.11$). On a single stream, EVIS achieves real-time generation at $30{\times}8$. With GPU-parallel batching, eight $60{\times}4$ streams remain real-time. Throughput per environment peaks at around 4 parallel environments.

\begin{table}[t]\centering\small\setlength{\tabcolsep}{6pt}
\caption{\textbf{Generation throughput on a single RTX 5090.}
For $1000{\times}1$+, we also render motion vectors and depth maps.
}
\begin{tabular}{ll c c c c}
\toprule
Method & Config & Envs & RTF & Wall\,(s) & Real-time\\
 & & & & \footnotesize per 10\,s data & \\
\midrule
\multirow{6}{*}{EVIS} & $1000{\times}1$ & 1 & 0.12 & 83.3 & \\
· & $1000{\times}1$+ & 1 & 0.11 & 90.9 & \\
 & ${125{\times}8}$ & 1 & {0.40} & {25.0} & \\
 & $30{\times}8$ & 1 & 1.20 & 8.3 & \cmark\\
& $60{\times}4$ & 4 & 1.02 & 9.8 & \cmark\\
 & $60{\times}4$ & 8 & 1.09 & 9.2 & \cmark\\
\cmidrule(lr){1-6}
\multirow{2}{*}{Vid2E} & $125{\times}8$ & 1 & 0.16 & 62.5 & \\
& $30{\times}8$ & 1 & 0.71 & 14.1 & \\
\bottomrule
\end{tabular}

\label{tab:benchmark}
\end{table}

\section{Conclusion}
We presented EVIS, a physics-grounded event data simulator that unifies linear-HDR event generation, closed-loop robot interaction, dense physical annotations, and GPU-parallel rollouts within a single real-time pipeline. By operating on the renderer's scene-linear HDR buffer and synthesizing intermediate frames through bidirectional motion-vector warping with depth-aware splatting, EVIS produces high-temporal-resolution events without the computational cost of dense rendering. 

Our experiments demonstrate that these design choices provide benefits beyond visual plausibility. Models trained exclusively on EVIS events transfer to real event-camera data, while pretrained models for reconstruction, feature matching, and point tracking operate on EVIS events without adaptation. Bidirectional motion warping outperforms video frame-interpolation methods based on pre-trained networks, while significantly reducing the computational cost. EVIS sustains real-time performance across multiple GPU-parallel environments.

Overall, these results demonstrate that EVIS is a scalable platform for generating realistic, action-conditioned events with synchronized physical supervision. We hope EVIS facilitates broader use of event cameras in robot perception by making large-scale, reproducible simulation practical for both training and evaluation.

{
    \small
    \bibliographystyle{ieeenat_fullname}
    \bibliography{main}
}
\appendix
\appendix

\twocolumn[{%
  \centering
  {\fontsize{12}{18}\selectfont\bfseries EVIS: Real-Time Event Camera Simulation with Multimodal Supervision in NVIDIA Isaac Sim Supplementary Material\par}
  \vspace{6mm}
}]

\section{Anti-aliasing}
\begin{figure}[h]
\centering
\includegraphics[width=0.7\linewidth]{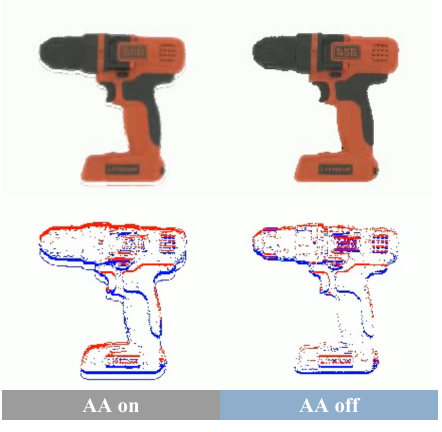}
\caption{\textbf{Artifacts caused by anti-aliasing.} Anti-aliasing produces near-zero motion vectors at softened edges, causing warping artifacts. Disabling it removes these artifacts.}
\vspace{-3mm}
\label{fig:aa}
\end{figure}

Isaac Sim enables anti-aliasing by default. While this is beneficial for full-resolution rendering, producing visually smoother and more natural edges, it is harmful to motion-vector warping. Anti-aliasing expands the effective footprint of an object by a thin fringe of blended pixels along its silhouette. However, the motion vectors provided by the renderer assign near-zero displacement to these fringe pixels rather than the true motion of the foreground object. These pixels therefore fail to track the object's movement during warping, leaving artifacts alongside object contours as shown in \cref{fig:aa}. To eliminate this issue, we disable anti-aliasing by default during rendering.

\section{Border handling}
Bidirectional warping relies on the assumption that every pixel in the target frame is visible in at least one of the two neighboring frames. However, pixels near image boundaries may have just entered or exited the field of view, leaving them without valid correspondences and introducing artifacts in the warped result. To address this issue, our method supports an optional over-rendering margin: during rendering, we extend each side of the frame by a configurable number of pixels and crop the excess after warping. This simple strategy produces event images free of visible boundary artifacts.

\section{Details of Sim-to-Real Experiment}
\paragraph{Dataset.}
Disc textures are generated procedurally. We first sample random white noise and apply Gaussian blurring to the resulting image. The blurred image is then thresholded at its median intensity to produce binary black-and-white patches. A gray reference wedge is placed at the top of each disc to serve as an absolute rotational angle reference. By varying the random seed and the Gaussian kernel standard deviation $\sigma$, we generate five distinct textures that differ in pattern shape and spatial density.

\paragraph{Ground Truth Rotation Speed.}
Ground-truth angular velocity for the real event clips is obtained in a training-free manner by exploiting the rotational periodicity of the event stream. Specifically, we bin events into 1 ms spatial descriptors, compute their Fast Fourier Transform (FFT) along the temporal axis, and identify the dominant peak in the resulting power spectrum as the revolution frequency  $f_{\text{rev}}$, yielding the angular velocity $\omega = 2\pi f_{\text{rev}}$.

\paragraph{Training Details.}
Each 0.5\,s event segment is represented as a per-polarity event voxel grid discretized into 128 temporal bins with linear interpolation applied along the time axis. Each bin is normalized by dividing by its maximum event count. We formulate rotational speed estimation as a regression task and train the network with the mean squared error (MSE) loss. Optimization is performed using AdamW with a weight decay of $1\times10^{-4}$ and an initial learning rate of $3\times10^{-4}$. The model is trained for 5 epochs.

\section{Experiment Results}
\begin{figure}[t]
\centering
\includegraphics[width=\linewidth]{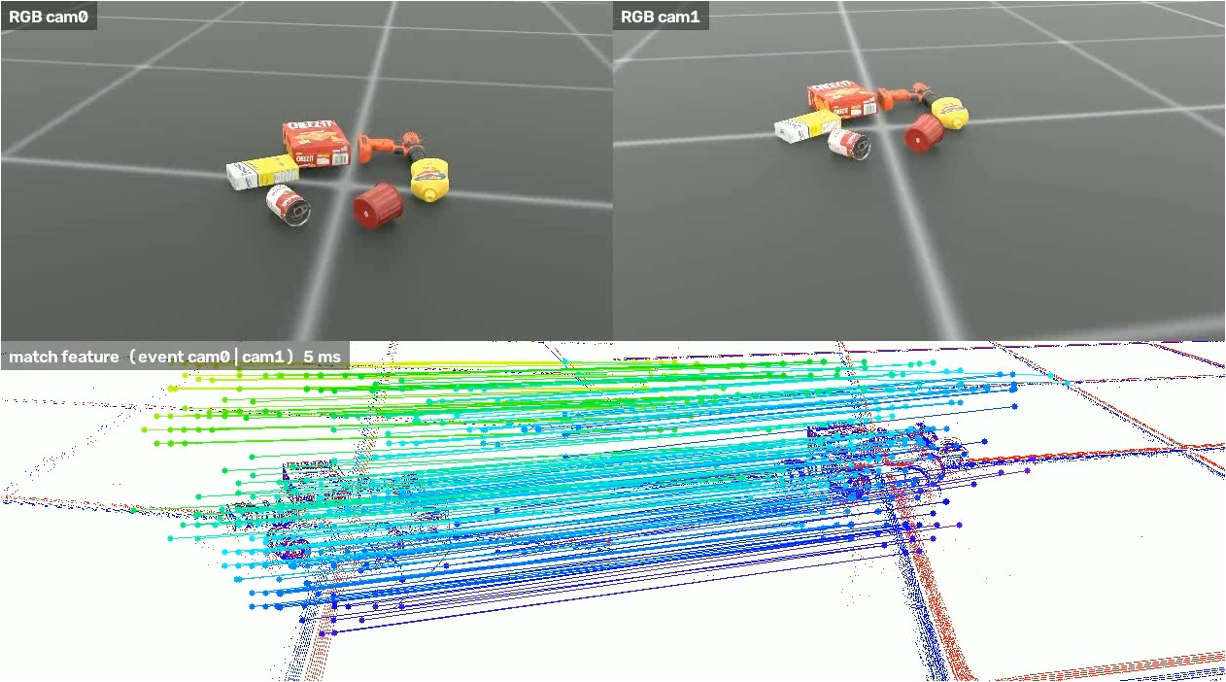}
\caption{{Event feature matching results in a stereo setup.} }

\label{fig:stereo}
\end{figure}

In addition to matching across viewpoints sampled along a single trajectory, we also evaluate matching performance under a stereo configuration using~\cite{zhang2026match} with a 128\,ms temporal window for inference.  \cref{fig:stereo} shows a representative matching result. We present additional qualitative examples across different tasks in the supplementary video.

\end{document}